\documentclass[journal]{IEEEtai}
\pdfoutput=1

\usepackage{amsthm}

\usepackage[colorlinks,urlcolor=blue,linkcolor=blue,citecolor=blue]{hyperref}
\usepackage{amsmath,amssymb}
\usepackage{graphicx}
\usepackage{booktabs}
\usepackage{multirow}
\usepackage{xcolor}
\usepackage{algorithm}
\usepackage{algorithmic}

\newcommand{\AP}{AP}

\begin{document}

\title{Neurosymbolic Routing for Reliable Reasoning on Resource-Constrained Edge Devices}

\author{Avyay Sadhu, Alvaro Velasquez, and Lekai Chen
\thanks{This work has been submitted to the IEEE for possible publication. Copyright may be transferred without notice, after which this version may no longer be accessible.}
\thanks{A. Sadhu is an independent researcher (e-mail: avyay.sadhu@gmail.com).}
\thanks{A. Velasquez and L. Chen are with the Department of Computer Science, University of Colorado Boulder, Boulder, CO 80309 USA (e-mail: alvaro.velasquez@colorado.edu; lekai.chen@colorado.edu).}}

\markboth{IEEE Transactions on Artificial Intelligence}
{Neurosymbolic Routing for Reliable Reasoning on Resource-Constrained Edge Devices}

\maketitle

\begin{abstract}
Running a language model on edge hardware provides private and low-latency reasoning without a network connection, and yet the small models that fit on such devices are unreliable on the tasks computers are expected to handle well, such as arithmetic, algebra, and formal logic problems. We argue that much of this unreliability is avoidable. Many queries appearing to demand reasoning are in fact structurally deterministic and permit fast and exact symbolic solutions. Therefore, forcing a probabilistic model to approximate them sacrifices accuracy and energy for little benefit. We present a neurosymbolic router that classifies each incoming query and dispatches it to the cheapest correct solver, sending structured tasks to deterministic engines and reserving the small language model (SLM) for open-ended word problems. Instead of hand-coding the routing logic, we learn a deterministic finite automaton (DFA) with the L* grammatical inference algorithm, using the SLM as a membership oracle and labeled data as an equivalence oracle. On a Raspberry Pi~4B (8\,GB RAM, no GPU), evaluated on 100 untested prompts from DeepMind Mathematics, GSM8K, and RuleTaker, learned routing attains 100\% routing accuracy and 98.3\% overall accuracy with a 512-token reasoning budget (93.3\% on word problems), compared with 72.0\% for the strongest agent baseline, Program-of-Thought, and 58.7\% for a tool-calling agent given the same solvers. Since formatted queries never reach the model, the router answers them in 1--11\,ms and, in its 30-token configuration, runs 8.8$\times$ faster and 2.8$\times$ more energy-efficient than Program-of-Thought.
\end{abstract}

\begin{IEEEImpStatement}
Edge devices such as Raspberry Pi computers increasingly host intelligent applications in healthcare, agriculture, and education, where cloud connectivity is often unavailable or undesirable for privacy reasons. The small language models that run on such hardware, however, are unreliable on structured reasoning, and the strongest small-model baseline we evaluated answered only 37\% of our benchmark correctly. Our router avoids this limitation by sending structured queries to purpose-built symbolic solvers and reserving the language model for genuinely open-ended problems. Because the routing logic is \emph{learned} through grammatical inference rather than hand-coded, it can be extended to new problem types from data alone. Compared with a tool-calling agent that has access to the same solvers, the learned router is both more accurate (78\% vs.\ 59\%) and 2.3$\times$ more energy-efficient on a Raspberry Pi~4B, and it answers structured queries in under 12 milliseconds. The architecture is model-agnostic and could be applied to offline tutoring systems, field calculators, or decision-support tools in which correctness and latency matter more than open-ended fluency.
\end{IEEEImpStatement}


% ======================================================================
\section{Introduction}
\label{sec:intro}
% ======================================================================

\IEEEPARstart{R}{unning} a language model entirely on a Raspberry Pi makes it possible to answer questions in a rural clinic with no internet, on a farm sensor beyond cellular range, or on a device that must never send a user's data off the premises. This is the appeal of edge inference, and it explains why interest in local deployment has grown in step with the latency, privacy, and connectivity constraints that often rule out the cloud~\cite{xu2024edgellm}. The appeal comes with a hard limit. A model small enough to fit in a few gigabytes of memory on a CPU-only board has modest capacity, and it struggles most on the questions that have a single correct answer, such as adding numbers, solving equations, and checking whether a set of facts entails a conclusion. In our experiments, the strongest such configuration answers only 37\% of a balanced reasoning benchmark correctly (Section~\ref{sec:versions}).

Throughout this paper, we use \textit{SLM} (small language model) for models with fewer than 10 billion parameters; our ``SLM-only baselines'' and ``neural inference'' all refer to this class.

Our approach begins from an observation that is easy to overlook. Many of the queries we ask language models to ``reason'' about are not open-ended at all. An arithmetic expression has exactly one correct value, a linear system has a closed-form solution, and a propositional-logic query is decidable by forward chaining. For problems of this kind, a deterministic solver is not merely competitive with a language model but exact, faster by orders of magnitude, and far cheaper in energy. Running an SLM on inputs whose structure is already known wastes computation and often introduces errors that a solver would not make. The difficulty is therefore not solving such problems but \emph{recognizing} them reliably before any expensive inference takes place.

\begin{figure*}[!t]
    \centering
    \includegraphics[width=0.95\textwidth]{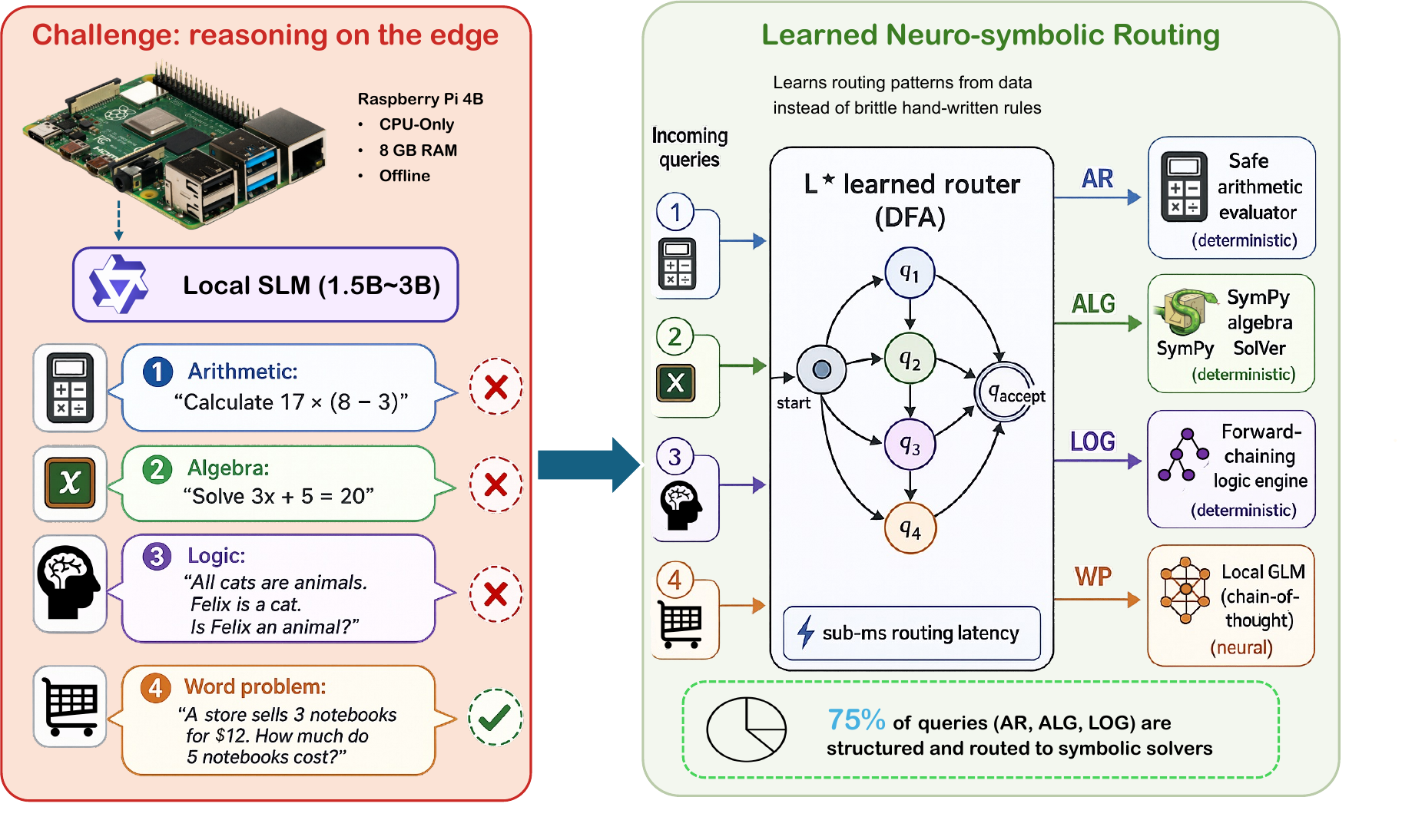}
    \caption{Overview of the proposed learned neurosymbolic routing architecture for reliable reasoning on resource-constrained edge devices. A local SLM running on a Raspberry Pi struggles with structured reasoning tasks, while an $L^{\ast}$-learned DFA router classifies incoming queries into arithmetic (AR), algebra (ALG), logic (LOG), or word-problem (WP) categories. Structured queries are dispatched to deterministic symbolic solvers, whereas open-ended word problems are routed to the local SLM, reducing unnecessary neural inference while preserving flexibility.}
    \label{fig:dfa-routing}
\end{figure*}

We propose a \textit{neurosymbolic adaptive router}, shown in Fig.~\ref{fig:dfa-routing}, that classifies each query into one of four categories, namely arithmetic (AR), algebra (ALG), formal logic (LOG), and word problems (WP), and dispatches it to the appropriate solver:
\begin{itemize}
    \item \textbf{AR} $\rightarrow$ safe arithmetic evaluator (Python \texttt{ast}),
    \item \textbf{ALG} $\rightarrow$ SymPy symbolic algebra solver,
    \item \textbf{LOG} $\rightarrow$ forward-chaining logic engine with closed-world assumption,
    \item \textbf{WP} $\rightarrow$ SLM with chain-of-thought prompting.
\end{itemize}

Of the four categories, only word problems, which demand natural-language comprehension and multi-step reasoning, genuinely require the language model. The three structured categories are handled by deterministic solvers that are exact on well-formed inputs and answer in milliseconds. In our balanced benchmark these account for three-quarters of all queries, so the model is invoked on only a minority of inputs; the exact fraction in deployment depends on the workload.

We evaluate this architecture on a Raspberry Pi~4B running Phi-4-mini (3.8B parameters, Q6\_K quantization) across 100 held-out prompts drawn from three established benchmarks (DeepMind Mathematics~\cite{saxton2019deepmind}, GSM8K~\cite{cobbe2021gsm8k}, and RuleTaker~\cite{clark2020ruletaker}) with three repeats per prompt (300 total trials). Our contributions are:

\begin{enumerate}
    \item \textbf{Learned routing via grammatical inference.} We apply the L* algorithm~\cite{Angluin87} to learn DFA-based query classifiers using the SLM as a membership oracle and ground-truth data as an equivalence oracle. This replaces brittle hand-coded routing rules with patterns learned from data, achieving 100\% classification accuracy on held-out prompts.
    \item \textbf{Fair baseline comparison.} We compare against a tool-calling agent that gives the same SLM access to the same symbolic solvers, isolating the routing mechanism as the independent variable. L*-learned routing achieves 78.0\% task accuracy vs.\ 58.7\% for the agent, while being 2.4$\times$ faster and 2.3$\times$ more energy-efficient.
    \item \textbf{Comparison with agent reasoning frameworks.} We evaluate three agent frameworks (ReAct~\cite{yao2023react}, Program-of-Thought~\cite{chen2023pot}, and Plan-and-Solve~\cite{wang2023plansolve}) on the same hardware and benchmark, demonstrating that L*-learned routing outperforms the best agent (Program-of-Thought, 72.0\%) by 6 percentage points while requiring 8.8$\times$ lower latency and 2.8$\times$ less energy per prompt.
    \item \textbf{System ablation.} An incremental ablation across five system versions demonstrates the marginal contribution of each symbolic solver and quantifies the accuracy--energy tradeoff on constrained hardware.
    \item Open-source release of the benchmark splits with full provenance, the learned routing automata, and per-trial result logs, together with a script that reproduces every reported number.\footnote{Anonymized repository for review: \url{https://anonymous.4open.science/r/NeurosymbolicRouting_Data-F701/}.}
\end{enumerate}

The remainder of this paper is organized as follows. Section~\ref{sec:related} reviews related work on edge inference, neurosymbolic systems, grammatical inference, and model routing. Section~\ref{sec:architecture} describes the router and its symbolic solvers, and Section~\ref{sec:gi} explains how the routing automaton is learned with L*. Section~\ref{sec:setup} details the experimental setup, Section~\ref{sec:versions} traces the incremental construction of the system, and Section~\ref{sec:results} reports the main results. Section~\ref{sec:analysis} examines why the approach works, Section~\ref{sec:discussion} discusses its limitations, and Section~\ref{sec:conclusion} concludes.

% ======================================================================
\section{Related Work}
\label{sec:related}
% ======================================================================

\subsection{Edge Inference}
Deploying language models on resource-constrained devices has been explored through quantization~\cite{dettmers2022gptint8, frantar2023gptq}, pruning~\cite{sun2024simpleprune}, and knowledge distillation~\cite{hinton2015distilling}. Frameworks such as llama.cpp~\cite{llamacpp2023} allow CPU-only inference for quantized models on commodity hardware. Our work is complementary, since we reduce the \textit{number} of queries routed to the SLM rather than optimizing the model itself.

\subsection{Neurosymbolic AI}
Combining neural and symbolic reasoning has a long history~\cite{garcez2019neuralsymbolic, kautz2022third}. Recent work integrates tool use into SLM pipelines. Toolformer~\cite{schick2023toolformer} teaches models to call APIs, while PAL~\cite{gao2023pal} generates executable code. These approaches rely on the model itself to decide when to invoke tools, which adds inference latency to every routing decision. Our approach differs in that routing decisions are made \textit{before} the SLM is invoked, using a learned DFA classifier with no inference cost.

\subsection{Agent Reasoning Framework Baselines}
\label{sec:agent_frameworks}

We additionally evaluate three agent reasoning frameworks on T2 (100 prompts, 1 trial each) to situate our approach within the broader landscape of tool-augmented inference. We select ReAct, Program-of-Thought, and Plan-and-Solve as representatives of the interleaved-reasoning, code-generation, and plan-then-execute families. Other frameworks follow related paradigms that we do not run. Reflexion~\cite{shinn2023reflexion} adds verbal self-reflection across multiple attempts, and CodeAct~\cite{wang2024codeact} frames actions as executable code, which places it close to Program-of-Thought. Because all systems use greedy decoding (temperature${}=0.0$, top-$k=1$, seed${}=42$), outputs are fully deterministic and a single trial per prompt is sufficient. These frameworks differ structurally from our approach, since they require one or more SLM inference steps \emph{per query} to decide which tool to invoke, whereas our learned DFA router makes routing decisions at sub-millisecond cost with no inference overhead on structured queries.

\textbf{ReAct}~\cite{yao2023react} interleaves Thought, Action, and Observation steps in a multi-turn loop (up to 6 steps). The model reasons about the problem, selects one of the four symbolic tools, observes the result, and repeats until it calls \texttt{finish()}.

\textbf{Program-of-Thought (PoT)}~\cite{chen2023pot} prompts the SLM to generate a Python program that solves the problem, which is then executed in a sandboxed environment with a restricted set of builtins. If execution fails, the model retries once with a simpler prompt.

\textbf{Plan-and-Solve}~\cite{wang2023plansolve} first generates an explicit numbered plan, then executes a single tool call based on the plan. We implement it with the same four symbolic tools as the tool-calling agent.

\subsection{Grammatical Inference}
Grammatical inference, the problem of learning formal languages from data, has a rich history in automata theory~\cite{gold1967language, higuera2010grammatical}. The L* algorithm~\cite{Angluin87} learns the minimal DFA for a regular language using membership and equivalence oracles in polynomial time and queries. RPNI~\cite{oncina1992rpni} takes a passive approach, learning from positive and negative examples without oracles. These methods have been applied to problems such as protocol inference~\cite{fiterau2020analysis} and model checking~\cite{peled2001model}. To our knowledge, this is the first application of grammatical inference to learn routing patterns for neurosymbolic query dispatch on edge devices.

\subsection{LLM Routing and Cascading}
FrugalGPT~\cite{chen2023frugalgpt} cascades queries through progressively larger models. Router-based systems like RouteLLM~\cite{ong2024routellm} learn to dispatch queries to strong or weak models. Our router operates at a different granularity. Rather than choosing between models of different sizes, it chooses between \textit{qualitatively different solvers}, namely symbolic engines and neural inference.

\subsection{Mathematical Reasoning}
Chain-of-thought prompting~\cite{wei2022cot} and its variants~\cite{wang2023selfconsistency} improve mathematical reasoning but remain probabilistic. Tool-augmented approaches~\cite{cobbe2021gsm8k} combine models with calculators. We extend this paradigm by routing entire categories to symbolic solvers, not just individual arithmetic operations within a generation.

% ======================================================================
\section{System Architecture}
\label{sec:architecture}
% ======================================================================

\subsection{Overview}

The system processes each query $q$ in two stages: (1)~a \textit{classifier} $C(q) \in \{\texttt{AR}, \texttt{ALG}, \texttt{LOG}, \texttt{WP}\}$ determines the problem category, and (2)~a \textit{dispatcher} routes $q$ to the corresponding solver. During development (V1--V5), the classifier is a hand-coded, regex-based function; in the final system it is replaced by a learned DFA with identical zero inference cost (Section~\ref{sec:gi}). The dispatcher invokes one of four solver modules.

\subsubsection{Solver Naming Convention}
Solvers are labeled A1--A6 following the project's internal development order, not the order of introduction in this paper. For clarity:
\begin{itemize}
    \item \textbf{A1}: Fast SLM inference (12 tokens, used as fallback).
    \item \textbf{A2}: Extended SLM inference (30--512 tokens, primary for WP).
    \item \textbf{A3}: Post-processing repair layer (re-prompts SLM with prior context).
    \item \textbf{A4}: SymPy algebra solver (deterministic, for ALG).
    \item \textbf{A5}: Safe arithmetic evaluator (deterministic, for AR).
    \item \textbf{A6}: Forward-chaining logic engine (deterministic, for LOG).
\end{itemize}

A1 and A3 serve as fallback and repair actions respectively; the primary solvers are A5 (AR), A4 (ALG), A6 (LOG), and A2 (WP).

\subsection{Primary Solvers}

\begin{itemize}
    \item \textbf{A5: Arithmetic evaluator.} Extracts numeric expressions from the query using regex, parses them into a Python AST, and evaluates using a whitelist of safe operators (\texttt{+}, \texttt{-}, \texttt{*}, \texttt{/}, \texttt{**}, \texttt{\%}, \texttt{//}). Only \texttt{ast.Constant} and \texttt{ast.BinOp} nodes are permitted; exponents are capped at 10{,}000.
    \item \textbf{A4: Algebra solver.} Extracts equations and target variables, inserts implicit multiplication (e.g., \texttt{3x} $\rightarrow$ \texttt{3*x}), and solves via \texttt{sympy.solve()}~\cite{sympy2023}. Supports single- and multi-variable linear systems.
    \item \textbf{A6: Logic engine.} Parses natural-language facts, rules, and queries from RuleTaker-style prompts into structured representations. Applies forward chaining with a closed-world assumption to derive new facts until a fixed point (bounded at 50 iterations), then evaluates the query against the derived fact set. See Section~\ref{sec:a6}.
    \item \textbf{A2: SLM chain-of-thought.} Constructs a system prompt (``Solve this math problem. Put your final answer in \texttt{\textbackslash boxed\{\}}''), sends the query to a locally served SLM via llama.cpp HTTP API, and extracts the final numeric answer.
\end{itemize}

\subsection{Classifier Design}

During development (V1--V5), the classifier used hand-coded regex rules applied in priority order (LOG $>$ ALG $>$ AR $>$ WP). While this achieved 100\% routing accuracy on our benchmark, such rules are brittle and do not generalize, since an adversarially phrased query can bypass pattern matching entirely.

In the final system, we replace the hand-coded classifier with a DFA learned via the L* algorithm (Section~\ref{sec:gi}). The learned classifier operates on the same priority ordering but derives its patterns from data rather than manual rules. It achieves 100\% routing accuracy on held-out test prompts with $<$\,0.05\,ms classification latency on the Pi, which is comparable to the regex classifier but with the advantage of being learned and thus adaptable to new domains.

\subsection{Fallback Chains}

When a primary solver fails (e.g., SymPy cannot parse an equation), the system escalates through a fallback chain:
\begin{itemize}
    \item AR: A5 $\rightarrow$ A1 $\rightarrow$ A2
    \item ALG: A4 $\rightarrow$ A1 $\rightarrow$ A2
    \item LOG: A6 $\rightarrow$ A1
    \item WP: A2 $\rightarrow$ A3 (repair)
\end{itemize}

\subsection{A6: Forward-Chaining Logic Engine}
\label{sec:a6}

The logic engine processes RuleTaker-style prompts through three stages:

\subsubsection{Parsing} Natural-language statements are converted to structured facts and rules:
\begin{itemize}
    \item ``Felix is a cat'' $\rightarrow$ \texttt{fact(felix, is, cat)}
    \item ``All cats are animals'' $\rightarrow$ \texttt{rule(?x, is, cat $\Rightarrow$ ?x, is, animal)}
    \item ``Is Felix an animal?'' $\rightarrow$ \texttt{query(felix, is, animal)}
\end{itemize}

\subsubsection{Forward Chaining} The engine iteratively applies rules to the current fact set, deriving new facts until no new facts can be produced (fixed-point convergence) or a safety bound of 50 iterations is reached. New facts are added eagerly (immediately available for subsequent rule applications within the same iteration).

\subsubsection{Query Evaluation} Under the closed-world assumption, a query is answered ``Yes'' if the queried fact exists in the derived set, and ``No'' otherwise.

% ======================================================================
\section{Learned Routing via Grammatical Inference}
\label{sec:gi}
% ======================================================================

A limitation of the hand-coded classifier described in Section~\ref{sec:architecture} is that its regex patterns are brittle, since a user may phrase a logic problem in a way that does not match the expected patterns, causing misrouting. We address this by \emph{learning} the routing classifier from data using grammatical inference.

\subsection{Token-Class Abstraction}
\label{sec:token_abstraction}

Raw query text has an unbounded vocabulary, making it unsuitable as input to a DFA. We define a finite alphabet of 12 abstract token classes:
\begin{center}
\small
\begin{tabular}{ll}
\textbf{Token} & \textbf{Matches} \\
\hline
\texttt{CMD} & calculate, solve, simplify, \ldots \\
\texttt{NUM} & numeric literals \\
\texttt{VAR} & single-letter variables ($x$, $y$, $m$) \\
\texttt{OP} & arithmetic operators ($+$, $-$, $\times$, $/$) \\
\texttt{PAREN} & parentheses and brackets \\
\texttt{ENT} & named entities (Felix, Allan, \ldots) \\
\texttt{PROP} & properties and adjectives (green, kind) \\
\texttt{RULE} & rule keywords (if, then, all, every) \\
\texttt{QMARK} & question indicators (is, does, ?) \\
\texttt{NARR} & narrative words (collapsed runs) \\
\texttt{UNIT} & units of measurement (\$, mph, kg) \\
\texttt{DOT} & sentence-ending punctuation \\
\end{tabular}
\end{center}

Each query is tokenized into a sequence over this 12-symbol alphabet, with consecutive \texttt{NARR} tokens collapsed and sequences truncated to 15 tokens. This abstraction preserves the structural signature of each category while reducing the input space to a finite alphabet suitable for DFA learning. Formally, we define the set of atomic propositions as $\AP = \{\texttt{CMD}, \texttt{NUM}, \texttt{VAR}, \texttt{OP}, \texttt{PAREN}, \texttt{ENT}, \texttt{PROP}, \texttt{RULE}, \texttt{QMARK}, \texttt{NARR}, \texttt{UNIT}, \texttt{DOT}\}$, giving $|\AP| = 12$. Since each token in a query is assigned to exactly one class, the alphabet used by $L^\star$ is the set of singleton labels $\Sigma = \{\{\ell\} : \ell \in \AP\}$, with $|\Sigma| = 12$. A query of length $n$ is thus represented as a word $w = \{\ell_1\}\{\ell_2\}\cdots\{\ell_n\} \in \Sigma^*$. The membership oracle $\mathcal{M}(w)$ operates on such words: given a token sequence $s$ (equivalently, a word $w \in \Sigma^*$), it returns whether $w$ belongs to the target category language $\mathcal{L}_c$.

\subsection{L* Algorithm for Route Learning}
\label{sec:lstar}

We adopt techniques from grammatical inference to guide our approach of learning the logic of routing as deterministic finite automaton (DFA). In particular, we leverage the $L^\star$ algorithm proposed in the seminal work of \cite{Angluin87}, which learns a regular language $\mathcal{L}$ through the use of a minimally adequate teacher that can answer two types of queries. When the learner executes a membership query, it presents a word $w \in \Sigma^*$ to the teacher and the teacher outputs whether $w \in \mathcal{L}$.
If the learner executes an equivalence query, it presents a hypothesis DFA to the teacher who must then answer whether this automaton encodes the language to be learned.
If not, the teacher generates a counterexample in the form of a word on which the two languages differ.
The $L^\star$ algorithm keeps track of state words $S \subseteq \Sigma^*$, which are closed under prefix operations (e.g. if $ab \in S$, then $a \in S$) and test words $E \subseteq \Sigma^*$, which are closed under suffix operations (e.g. if $ab \in E$, then $b \in E$).
Initially, we have $S = E = \{\varepsilon\}$, where $\varepsilon$ is the empty string.
As the algorithm proceeds, there are two critical properties that must be tracked and revolve around the notion of $E$-equivalence.

\newtheorem{definition}{Definition}[section] % Defines the 'definition' environment, numbered by section

\begin{definition}[$E$-Equivalence]
Given $w, w' \in \Sigma^*$ and set $E \subseteq \Sigma^*$, the words $w$ and $w'$ are $E$-equivalent with respect to the language $\mathcal{L}$, denoted $w \equiv_E w'$, if $w  e \in \mathcal{L} \iff w'  e \in \mathcal{L}$ holds for every $e \in E$.
\end{definition}

%\begin{definition}[Consistency]
%Given $(S, E)$, the consistency property holds if and only if there are no two state words in $S$ that are $E$-equivalent. That is $\forall s \in S, \nexists s' \in S, s \equiv_E s'$.
%\end{definition}

\begin{definition}[Consistency]
Given $(S, E)$, the consistency property holds if and only if for all $s, s' \in S$, the following implication holds: if $s \equiv_E s'$, then $s \ell \equiv_E s' \ell$ for all $\ell \in \Sigma$.

\end{definition}

\begin{definition}[Closedness]
Given $(S, E)$, the closedness property holds if and only if for all $s \in S$ and $\ell \in \Sigma$, there exists some $s' \in S$ such that $s  \ell \equiv_E s'$.
\end{definition}

For a closed and consistent $(S, E)$, a corresponding automaton can be derived by taking each $E$-equivalence class of $S$ to be a state, with the empty string $\varepsilon$ as the starting state.
The transition function is defined using the closedness property. That is, whenever we have $s  \ell \equiv_E s'$ per the closedness property, then we also have the transition from $s$ to $s'$ upon observing the label $\ell$ in the DFA.
Furthermore, it follows from the consistency property that $s'$ is unique.
The accepting states are those in the language of the teacher.

\begin{table}[t]
\centering
\caption{Example observation table $(S, E, T)$ built by $L^\star$ for $\mathcal{L}_{AR}$. $\checkmark$ = word belongs to the language, $\times$ = rejected. Rows above the dashed line are state words $S$; rows below are frontier words $S \cdot \Sigma$.}
\label{tab:lstar_table}
\small
\begin{tabular}{l|ccc}
\hline
$S \setminus E$ & $\varepsilon$ & \texttt{CMD} & \texttt{CMD{\tiny·}NUM} \\
\hline
$\varepsilon$                    & $\times$ & $\times$ & $\times$ \\
\texttt{CMD}                     & $\times$ & $\times$ & $\times$ \\
\texttt{CMD{\tiny·}NUM}          & $\times$ & $\checkmark$ & $\checkmark$ \\
\hline
\texttt{CMD{\tiny·}NUM{\tiny·}OP}     & $\times$ & $\times$ & $\checkmark$ \\
\texttt{CMD{\tiny·}NUM{\tiny·}OP{\tiny·}NUM} & $\times$ & $\checkmark$ & $\checkmark$ \\
\hline
\end{tabular}
\end{table}

\begin{figure}[t]
    \centering
    \includegraphics[width=\columnwidth]{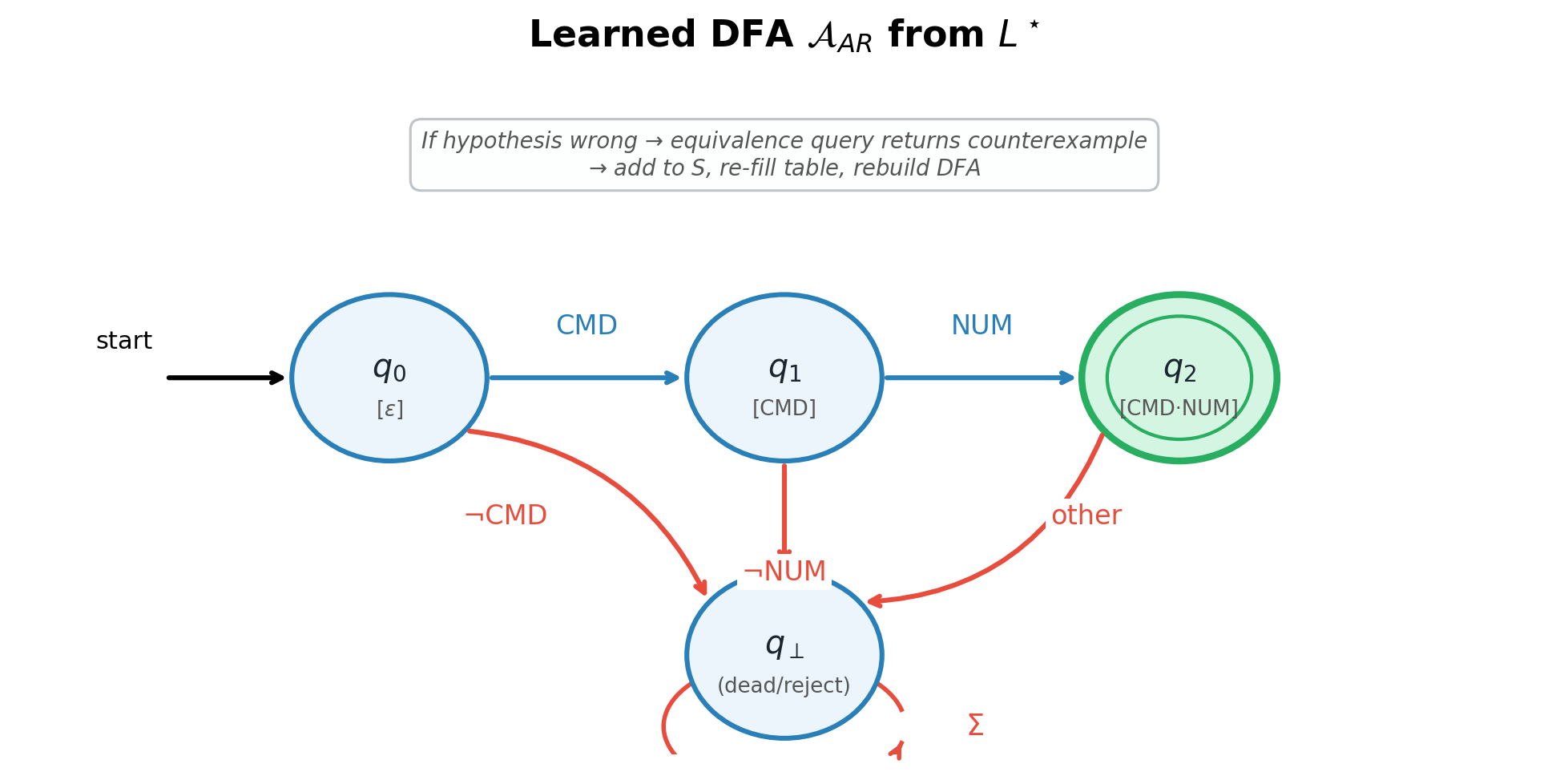}
    \caption{DFA $\mathcal{A}_{AR}$ learned by $L^\star$ from the observation table above. States correspond to $E$-equivalence classes of $S$; $q_2$ (double circle) is the accepting state. Red transitions lead to the dead/reject state $q_\perp$.}
    \label{fig:lstar_illustration}
\end{figure}

\paragraph{Running example.}
Consider learning $\mathcal{L}_{AR}$, the language of arithmetic queries.
$L^\star$ begins with $S = E = \{\varepsilon\}$ and an empty observation table.
It first asks: does $\{\texttt{CMD}\}$ belong to $\mathcal{L}_{AR}$?
The SLM oracle answers No (a bare command word is not an arithmetic query).
It then asks: does $\{\texttt{CMD}\}\{\texttt{NUM}\}$ belong to $\mathcal{L}_{AR}$?
Again No, a command followed by a single number is incomplete.
After further membership queries, $L^\star$ discovers that
$\{\texttt{CMD}\}\{\texttt{NUM}\}\{\texttt{OP}\}\{\texttt{NUM}\}$
(e.g., ``calculate 4 + 3'') is accepted.
The resulting observation table (Table~\ref{tab:lstar_table}) is closed and consistent,
so $L^\star$ constructs a hypothesis DFA and submits an equivalence query.
The equivalence oracle (tested against T1) finds no counterexample,
confirming the hypothesis. The learned DFA $\mathcal{A}_{AR}$ is shown in
Fig.~\ref{fig:lstar_illustration}: $q_0$ is the start state, $q_2$ accepts
sequences matching the arithmetic pattern, and $q_\perp$ rejects all others.
At query time, a new prompt is tokenized and run through this DFA in
$<$\,0.05\,ms with no SLM call required.

The $L^\star$ algorithm ensures that $(S, E)$ remains closed and consistent until the target language is learned. This is done by constructing a DFA from $(S, E)$.
If a counterexample is produced, it is used to modify $(S, E)$ by adding the counterexample and all its prefixes to $S$. Membership queries are then used to determine the values of the new entries.

We learn a one-vs-rest binary DFA for each category $c \in \{\texttt{AR}, \texttt{ALG}, \texttt{WP}, \texttt{LOG}\}$ using the L* algorithm~\cite{Angluin87}. L* maintains an observation table and iteratively refines a hypothesis DFA through two oracle queries:

\begin{itemize}
    \item \textbf{Membership oracle} $\mathcal{M}(s) \to \{0,1\}$: Given a token sequence $s$, returns whether $s$ belongs to category $c$. We implement this using the SLM itself: the query ``Is the following problem [category description]?'' is sent to Phi-4-mini, which responds Yes or No. For synthetic sequences generated during L* exploration (which have no natural-language counterpart), we fall back to a feature-based classifier.
    \item \textbf{Equivalence oracle} $\mathcal{E}(H) \to \{s \mid s \text{ is counterexample}\}$: Given a hypothesis DFA $H$, returns a counterexample where $H$ disagrees with ground-truth labels, or confirms that $H$ is correct. We implement this by testing $H$ against the labeled training set (T1, 40 prompts).
\end{itemize}

L* converges in polynomial time in the number of states of the target DFA, producing the minimal DFA consistent with the oracle responses.

\subsection{RPNI: Passive DFA Learning as a Component Variant}
\label{sec:rpni}

RPNI (Regular Positive and Negative Inference)~\cite{oncina1992rpni} is a passive grammatical inference algorithm that requires no oracle. RPNI builds a prefix tree acceptor from positive examples and iteratively merges states, rejecting any merge that would accept a negative example. In our system, RPNI serves as a \emph{replaceable component} that can substitute for L* as the DFA learning module without changing the routing architecture, dispatcher, or solvers. This comparison isolates the contribution of the SLM membership oracle, the novel element of our approach, to overall classification and task accuracy.

\subsection{Multi-Class Classification}
\label{sec:multiclass}

The four binary DFAs are composed into a multi-class classifier with priority ordering: \texttt{LOG} $>$ \texttt{ALG} $>$ \texttt{AR} $>$ \texttt{WP}. A query is classified by the highest-priority DFA that accepts it; if no DFA accepts, it defaults to \texttt{WP} (the most common fallback).

% ======================================================================
\section{Experimental Setup}
\label{sec:setup}
% ======================================================================

\subsection{Hardware}

All experiments run on a single Raspberry Pi~4 Model B with 8\,GB RAM, a quad-core ARM Cortex-A72 CPU (1.8\,GHz), and no GPU or neural accelerator. The operating system is Raspberry Pi OS (64-bit, Debian-based). Models are served via llama.cpp~\cite{llamacpp2023} in CPU-only mode with 4 threads.

\subsection{Models}

We evaluate two quantized SLMs:
\begin{itemize}
    \item \textbf{Phi-4-mini}~\cite{abdin2024phi4} (Microsoft, 3.8B parameters, Q6\_K quantization, 3.16\,GB)
    \item \textbf{Qwen2.5-Math-1.5B}~\cite{qwen2025} (Alibaba, 1.5B parameters, Q6\_K quantization)
\end{itemize}

All SLM calls use greedy decoding (temperature${}=0.0$, seed${}=42$), stopping at the first newline.

\subsection{Dataset}
\label{sec:dataset}

The evaluation benchmark comprises 100 held-out test prompts (25 per category) sampled from three established datasets, with full provenance tracking (source dataset, record ID, and field mapping version recorded for every prompt):
\begin{itemize}
    \item \textbf{AR} (25 prompts): From the \textit{DeepMind Mathematics Dataset}~\cite{saxton2019deepmind}, modules including addition/subtraction, multiplication, division, and mixed operations.
    \item \textbf{ALG} (25 prompts): From the \textit{DeepMind Mathematics Dataset}~\cite{saxton2019deepmind}, modules including linear 1D, linear 2D, and polynomial roots.
    \item \textbf{LOG} (25 prompts): From \textit{RuleTaker}~\cite{clark2020ruletaker}, forward-chaining logical entailment queries over natural-language facts and rules.
    \item \textbf{WP} (25 prompts): From \textit{GSM8K}~\cite{cobbe2021gsm8k}, multi-step grade-school word problems requiring natural-language comprehension.
\end{itemize}

Prompts were sampled with a fixed global seed (42) using stratified random sampling from the original dataset splits. No prompts were paraphrased, generated, or modified beyond whitespace normalization. Each prompt's category label is determined by the source dataset it was drawn from: DeepMind Mathematics arithmetic modules map to AR, algebra modules to ALG, RuleTaker queries to LOG, and GSM8K problems to WP.

The prompts are divided into two disjoint tiers, verified by both source key and text hash:
\begin{itemize}
    \item \textbf{T1} (40 prompts, 10 per category): Used for DFA training (L* equivalence oracle, RPNI examples).
    \item \textbf{T2} (100 prompts, 25 per category): Held-out evaluation set. All reported results use T2.
\end{itemize}

This train/test split ensures that routing accuracy is measured on prompts \emph{never seen during DFA learning}, addressing potential overfitting concerns. The benchmark is intentionally small (100 test prompts, 300 trials) due to the slow inference speed on the Pi (approximately 2 minutes per word problem at 512 tokens). We acknowledge this limits statistical power; the 3 repeats are primarily useful for measuring SLM variance on WP, as the symbolic solvers are deterministic. Larger-scale evaluation is an important direction for future work.

Each prompt is run 3 times per system configuration, yielding 300 trials per system. Evaluation uses \textit{exact match}, meaning the extracted answer must equal the ground truth.

\subsection{Tool-Calling Agent Baseline}
\label{sec:agent_baseline}

To provide a fair comparison, we implement a tool-calling agent that gives the same SLM (Phi-4-mini) access to the same four symbolic solvers via a function-calling prompt. The system prompt describes the available tools:
\begin{itemize}
    \item \texttt{arithmetic\_eval(expr)}: Safe arithmetic evaluator
    \item \texttt{sympy\_solve(eq, var)}: SymPy algebra solver
    \item \texttt{logic\_engine(answer)}: Logic inference
    \item \texttt{direct\_answer(number)}: Direct numeric response
\end{itemize}

The SLM receives each query along with tool descriptions and few-shot examples, then generates a single tool call. The output is parsed and the corresponding tool is executed. This mirrors the Toolformer~\cite{schick2023toolformer} paradigm, in which the model decides which tool to use at inference time. Unlike our learned router, the agent requires a full SLM inference step for \emph{every} query, even simple arithmetic that the symbolic solver could handle in 1\,ms.

\subsection{Metrics}

\begin{itemize}
    \item \textbf{Accuracy}: Fraction of trials with exact-match correct answers.
    \item \textbf{Latency}: Wall-clock time per inference, measured from prompt submission to answer extraction (millisecond precision).
    \item \textbf{Energy}: Total run energy measured via USB power meter (start/end mWh readings per run), divided by number of prompts.
\end{itemize}

\subsection{Error Taxonomy}

We classify errors by failure mode: \textbf{E0} correct; \textbf{E1} arithmetic error; \textbf{E2} logic error; \textbf{E3} algebra error; \textbf{E5} word problem error; \textbf{E7} timeout; \textbf{E8} parse failure.

% ======================================================================
\section{System Versions and Ablation}
\label{sec:versions}
% ======================================================================

Before learning the router, we built the symbolic backend incrementally across five versions, each adding or modifying exactly one component. Because the versions differ by a single change, the progression also serves as an ablation in which the accuracy and energy differences between successive versions isolate the contribution of each solver. We summarize the trajectory below, and Table~\ref{tab:main_results} reports the corresponding numbers.

\subsection{Baseline: SLM-Only}

The natural starting point is the SLM alone, with no routing and no symbolic solvers. We swept four configurations that vary two factors. The token budget is either 12 tokens (A1) or 30 tokens (A2), and decoding is either grammar-constrained, using a BNF grammar that fixes the output format, or free-form. Even the best of the four, namely 30 tokens with grammar constraints on Phi-4-mini, reaches only 37.3\% overall (48\% AR, 40\% ALG, 53.3\% LOG, 8\% WP), and 17\% of all trials fail simply because the answer cannot be parsed from the model's output. This is the gap that the rest of the system closes.

\subsection{V1: Arithmetic Solver (+A5)}

Routing AR queries to the safe arithmetic evaluator produces the first and largest jump: AR accuracy climbs from 48\% to a perfect 100\%, its latency collapses from 6.2\,s to 1\,ms, and overall accuracy rises to 53.0\%. Replacing approximation with exact computation, in one category, recovers more than fifteen points.

\subsection{V2: Algebra Solver (+A4)}

Adding the SymPy-based algebra solver for ALG repeats the pattern: ALG accuracy rises from 40\% to 96--100\% and its latency drops from 5.8\,s to 11\,ms, lifting the system to 67.0\% overall.

\subsection{V3: Post-Processing Repair Layer (+A3)}

V3 introduces an answer-repair module (A3) that extracts and corrects answers from verbose SLM output. Applied to word problems it helps modestly, raising WP from 8\% to 18.7\%, but our first attempt also applied it to logic queries, where it backfired. A repair layer built to recover numbers corrupts yes/no answers by pulling spurious figures out of the reasoning text, and LOG fell from 64\% to 36.7\%. Restricting A3 to numeric categories restores LOG to 64\% and brings the system to 70.7\%. The lesson is narrow but practical, namely that post-processing is format-sensitive and should not be applied across output types without validation, and we report it because it shaped the final design.

\subsection{V4: Confidence Calibration}

V4 adds a confidence calibrator, and its result is instructive precisely because it is null, with accuracy unchanged at 70.7\%. The calibrator flags low-confidence answers accurately, but flagging a likely-wrong answer does not make it correct when the model simply lacks the tokens to reason through the problem. The bottleneck, it turns out, is the reasoning budget itself rather than the system's ability to detect failure.

\subsection{V5: Logic Engine and a Larger Reasoning Budget (+A6, +CoT)}

V5 addresses the two remaining bottlenecks directly. First, a forward-chaining logic engine (A6) replaces SLM inference for LOG, reaching 100\% accuracy in under 1\,ms. Second, word problems, the one category that genuinely needs the model, are given a 512-token chain-of-thought budget in place of the 30-token cap, prompted to place the final answer in \texttt{\textbackslash boxed\{\}}. Together these lift the system to 99.0\% overall: AR, ALG, and LOG are perfect, and WP reaches 96\%. A single word problem fails consistently across all three repeats; we dissect it in Section~\ref{sec:error_analysis}.

% ======================================================================
\section{Results}
\label{sec:results}
% ======================================================================

\subsection{Learned Routing vs.\ Tool-Calling Agent}

Table~\ref{tab:gi_results} presents our central comparison, in which learned routing is set against a tool-calling agent given the same SLM and the same symbolic solvers, so that the routing mechanism is the only variable. All systems run Phi-4-mini on identical hardware with identical decoding parameters, end-to-end on T2 (100 prompts, 3 repeats, 300 trials).

At its full 512-token reasoning budget, the L*-learned router answers \textbf{98.3\%} of trials correctly (100\% on arithmetic, algebra, and logic, and 93.3\% on word problems), compared with 58.7\% for the tool-calling agent, a difference of close to forty points. The agent fails where it must both decide to call a tool and call it correctly. It falls to 59\% on logic, where it frequently never invokes the engine, and to 76\% on algebra, where it emits malformed tool calls. Because routing is settled in advance by the DFA, the learned router never faces this decision at inference time, and it dispatches all three structured categories with perfect accuracy.

The router is also the more efficient system. Structured queries, which make up three-quarters of the benchmark, are answered by symbolic solvers in 1--11\,ms with no model call at all, so at its 30-token operating point the router runs 2.4$\times$ faster than the agent (6{,}907\,ms vs.\ 16{,}439\,ms mean latency) and draws 2.3$\times$ less energy (33.1 vs.\ 75.0\,mWh per prompt). Raising the word-problem budget to 512 tokens improves accuracy at a higher cost on that one category, 141.3\,mWh per prompt overall, a trade-off we return to in Section~\ref{sec:discussion}.

What separates the two operating points is the token budget rather than the routing. The same learned router scores 78.0\% at 30 tokens and 98.3\% at 512 tokens, while its routing accuracy remains at 100\% throughout, and the gap to the hand-coded V5 system (99.0\%) nearly closes. Learning the router therefore costs essentially nothing in end-to-end accuracy.

\begin{figure}[t]
    \centering
    \includegraphics[width=\columnwidth]{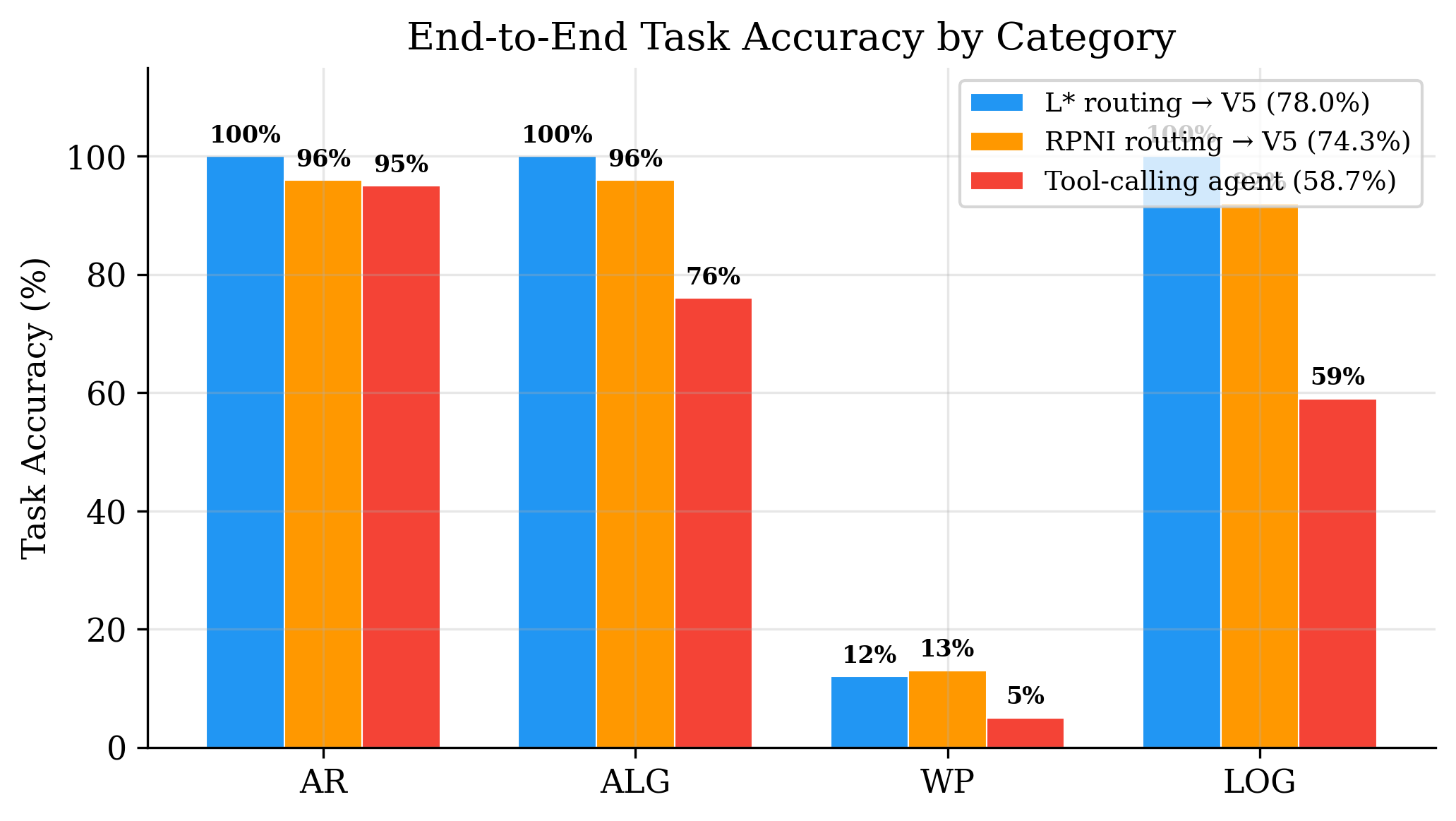}
    \caption{End-to-end task accuracy by category. L*-learned routing achieves 100\% on AR, ALG, and LOG via symbolic solvers, while the tool-calling agent struggles on all categories, particularly LOG (59\%) and ALG (76\%). Word problems remain challenging for all systems.}
    \label{fig:e2e_comparison}
\end{figure}

\begin{figure}[t]
    \centering
    \includegraphics[width=\columnwidth]{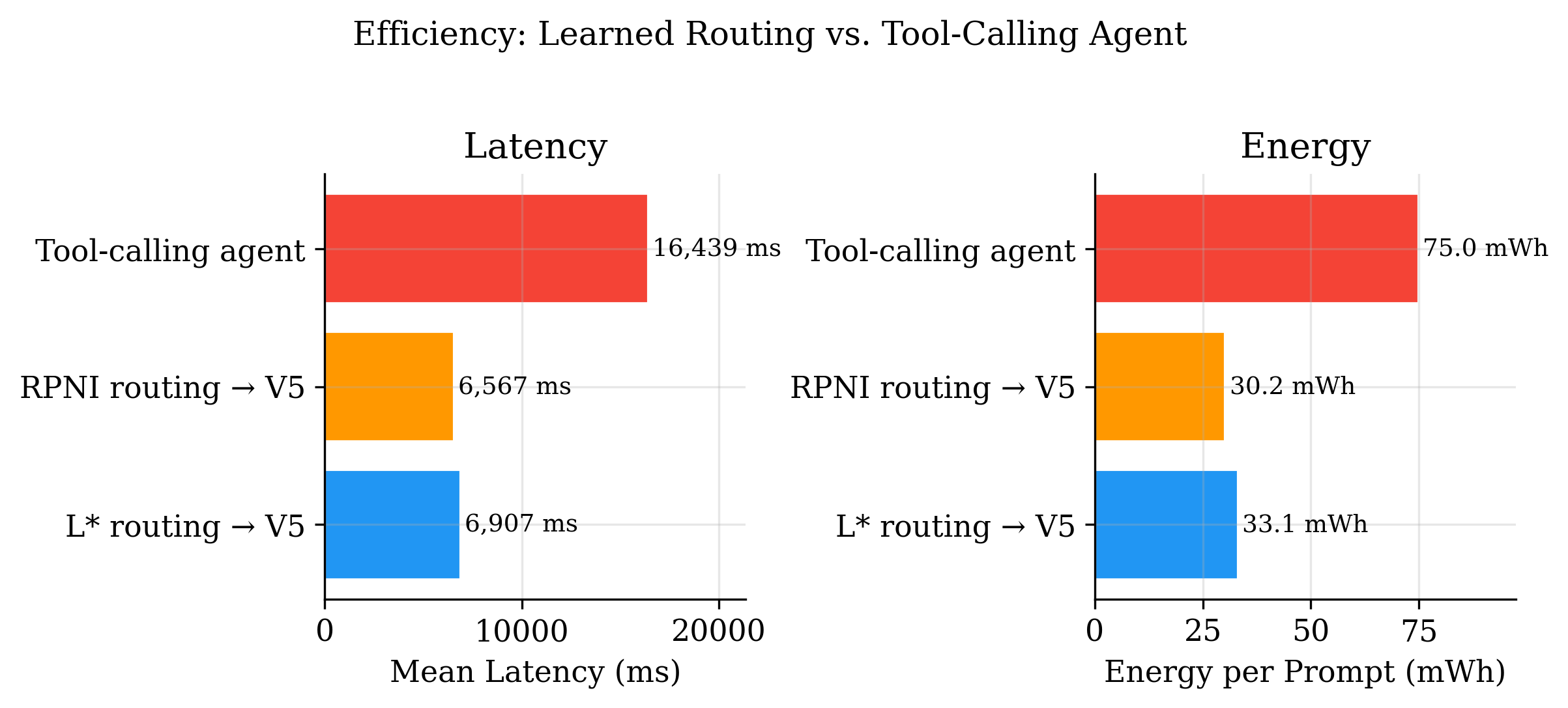}
    \caption{Efficiency comparison. Learned routing (L* and RPNI) is 2.4$\times$ faster and 2.3$\times$ more energy-efficient than the tool-calling agent, because AR, ALG, and LOG queries bypass the SLM entirely.}
    \label{fig:efficiency}
\end{figure}

RPNI-learned routing (74.3\%) also outperforms the agent despite using no oracle during training, demonstrating that even passive grammatical inference produces effective routing. The 3.7 percentage-point gap between L* and RPNI reflects the value of the SLM membership oracle: L* achieves 100\% routing accuracy on T2 versus 86\% for RPNI (which misclassifies 9 of 25 word problems due to their linguistic variety). Fig.~\ref{fig:e2e_comparison} visualizes the per-category comparison, and Fig.~\ref{fig:efficiency} shows the efficiency gap.

\subsection{Comparison with Agent Reasoning Frameworks}

Table~\ref{tab:gi_results} also includes three agent reasoning baselines evaluated on the same hardware. Program-of-Thought achieves the highest accuracy among agents (72.0\%), but still trails L*-learned routing by 6 percentage points. PoT's strength is word problems (88\%) and algebra (88\%), where generating executable Python code is effective, but it fails on logic problems (16\%), as the model struggles to translate natural-language rules into correct Python. ReAct achieves 63.0\% overall, performing reasonably on WP (68\%) but poorly on ALG (52\%) due to tool-call format errors when equations become complex. Plan-and-Solve performs worst at 43.0\%, suggesting that Phi-4-mini (3.8B parameters) lacks the capacity to reliably follow the two-stage plan-then-execute format.

The efficiency gap between agent frameworks and learned routing is substantial (Table~\ref{tab:efficiency}). ReAct averages 2.43 SLM calls per query, a multi-turn loop that compounds latency and energy, reaching 97.2\,s and 150.3\,mWh per prompt. L*-learned routing, by contrast, routes 75\% of queries to symbolic solvers in under 12\,ms with no SLM call at all. This structural advantage, routing \emph{before} inference rather than \emph{during} it, is the key differentiator.

\begin{table}[t]
\centering
\caption{End-to-End Task Accuracy on T2 (300 trials, 3 repeats). The 512-token result uses chain-of-thought prompting for WP; agent baselines use 1 repeat.}
\label{tab:gi_results}
\begin{tabular}{lccccc}
\hline
\textbf{System} & \textbf{AR} & \textbf{ALG} & \textbf{WP} & \textbf{LOG} & \textbf{All} \\
\hline
\textbf{L* routing $\to$ V5 (512tok)} & \textbf{100} & \textbf{100} & \textbf{93} & \textbf{100} & \textbf{98.3} \\
L* routing $\to$ V5 (30tok) & 100 & 100 & 12 & 100 & 78.0 \\
RPNI routing $\to$ V5 & 96 & 96 & 13 & 92 & 74.3 \\
\hline
\multicolumn{6}{l}{\textit{Agent framework baselines}} \\
Program-of-Thought~\cite{chen2023pot} & 96 & 88 & 88 & 16 & 72.0 \\
ReAct~\cite{yao2023react} & 92 & 52 & 68 & 40 & 63.0 \\
Tool-calling agent & 95 & 76 & 5 & 59 & 58.7 \\
Plan-and-Solve~\cite{wang2023plansolve} & 52 & 44 & 20 & 56 & 43.0 \\
\hline
\end{tabular}
\end{table}

\begin{table}[t]
\centering
\caption{Efficiency Comparison on T2 (Same Hardware, Same Model)}
\label{tab:efficiency}
\begin{tabular}{lrrr}
\hline
\textbf{System} & \textbf{Lat.\ (ms)} & \textbf{mWh/p} & \textbf{Route (\%)} \\
\hline
L* $\to$ V5 & 6,907 & 33.1 & 100.0 \\
RPNI $\to$ V5 & 6,567 & 30.2 & 86.0 \\
\hline
Program-of-Thought & 60,870 & 94.7 & n/a \\
ReAct & 97,193 & 150.3 & n/a \\
Tool-calling agent & 16,439 & 75.0 & n/a \\
Plan-and-Solve & 82,521 & 123.9 & n/a \\
\hline
\end{tabular}
\end{table}

\subsection{Classification Accuracy}

Table~\ref{tab:classify} shows routing classification accuracy (independent of task solving) for the learned classifiers. L* with both the SLM oracle and the feature-based oracle achieve 100\% on held-out T2, while RPNI achieves 86\%, primarily due to word-problem misclassification (64\% WP routing accuracy).

\begin{table}[t]
\centering
\caption{Routing Classification Accuracy on T2 (100 Prompts)}
\label{tab:classify}
\begin{tabular}{lccccc}
\hline
\textbf{Classifier} & \textbf{AR} & \textbf{ALG} & \textbf{WP} & \textbf{LOG} & \textbf{All} \\
\hline
L* (SLM oracle) & 100 & 100 & 100 & 100 & 100.0 \\
L* (feature oracle) & 100 & 100 & 100 & 100 & 100.0 \\
RPNI (passive) & 96 & 92 & 64 & 92 & 86.0 \\
Feature classifier & 100 & 100 & 100 & 100 & 100.0 \\
\hline
\end{tabular}
\end{table}

\subsection{V1--V5 Development Ablation}

% OLD TABLE (preserved for reference)
% \begin{table*}[t]
% \centering
% \caption{Main Results: V1--V5 Development (300 Trials Per System)}
% \label{tab:main_results_old}
% \begin{tabular}{lcccc}
% \hline
% \textbf{System} & \textbf{Acc.\ (\%)} & \textbf{Med.\ Latency (ms)} & \textbf{Energy (mWh/p)} & \textbf{Parse Fail (\%)} \\
% \hline
% Baseline (C2, grammar) & 37.3 & 8,612 & 75.0 & 17.0 \\
% Baseline (C2, no grammar) & 37.0 & 8,709 & 64.6 & 17.0 \\
% Baseline (C1, grammar) & 36.0 & 6,021 & 32.1 & 26.0 \\
% Baseline (C1, no grammar) & 33.0 & 9,080 & 59.9 & 26.0 \\
% \hline
% Hybrid V1 (+A5) & 53.0 & 4,147 & 75.9 & 15.0 \\
% Hybrid V2 (+A4) & 67.0 & 2,598 & 37.2 & 15.0 \\
% Hybrid V3.1 (+A3) & 70.7 & 1,413 & 34.0 & 4.3 \\
% Hybrid V4 (+Calibration) & 70.7 & 1,466 & 34.0 & 4.0 \\
% \hline
% V5 Default (30 tokens) & 79.7 & 6 & 30.6 & 4.0 \\
% V5 (Phi, 300 tokens) & 98.0 & 6 & 147.2 & 0.0 \\
% V5 (Phi, 512 tokens) & 99.0 & 6 & 211.1 & 0.0 \\
% V5 (Qwen, 512 tokens) & 99.0 & 6 & 147.6 & 0.0 \\
% \hline
% \end{tabular}
% \end{table*}

Table~\ref{tab:main_results} summarizes the incremental development from SLM-only baselines through V5. Because each version adds a single component, the V1--V5 progression serves as a natural ablation that isolates the contribution of that component. The best system with hand-coded routing (V5, Phi-4-mini, 512 tokens) reaches 99.0\% accuracy, while the 30-token default configuration reaches 79.7\%.

\begin{table}[t]
\centering
\caption{V1--V5 Development Ablation (300 Trials Each, Hand-Coded Routing)}
\label{tab:main_results}
\begin{tabular}{lccc}
\hline
\textbf{System} & \textbf{Acc.\ (\%)} & \textbf{Med.\ Lat.\ (ms)} & \textbf{E (mWh/p)} \\
\hline
SLM-only baseline & 37.3 & 8,612 & 75.0 \\
V1 (+A5 arith.) & 53.0 & 4,147 & 75.9 \\
V2 (+A4 algebra) & 67.0 & 2,598 & 37.2 \\
V3.1 (+A3 repair) & 70.7 & 1,413 & 34.0 \\
V5 (30 tokens) & 79.7 & 6 & 30.6 \\
V5 (512 tokens) & 99.0 & 6 & 211.1 \\
\hline
\end{tabular}
\end{table}

\subsection{Per-Category Accuracy}

Table~\ref{tab:per_category} shows accuracy by problem category for both the learned-routing systems and the V1--V5 ablation. Symbolic solvers achieve perfect accuracy on their respective domains.

% OLD per-category table (preserved for reference)
% \begin{table}[t]
% \centering
% \caption{Per-Category Accuracy (\%) Across Key Systems (Old)}
% \begin{tabular}{lccccc}
% \hline
% \textbf{System} & \textbf{AR} & \textbf{ALG} & \textbf{LOG} & \textbf{WP} & \textbf{All} \\
% \hline
% Baseline (best) & 48.0 & 40.0 & 53.3 & 8.0 & 37.3 \\
% V1 (+A5) & 100 & 40.0 & 64.0 & 8.0 & 53.0 \\
% V2 (+A4) & 100 & 96.0 & 64.0 & 8.0 & 67.0 \\
% V3.1 (+A3) & 100 & 100 & 64.0 & 18.7 & 70.7 \\
% V5 Default & 100 & 100 & 100 & 18.7 & 79.7 \\
% V5 (512tok) & 100 & 100 & 100 & 96.0 & 99.0 \\
% \hline
% \end{tabular}
% \end{table}

\begin{table}[t]
\centering
\caption{Per-Category Accuracy (\%): Learned Routing vs.\ Development Ablation}
\label{tab:per_category}
\begin{tabular}{lccccc}
\hline
\textbf{System} & \textbf{AR} & \textbf{ALG} & \textbf{WP} & \textbf{LOG} & \textbf{All} \\
\hline
\multicolumn{6}{l}{\textit{Learned routing (this work)}} \\
\textbf{L* routing $\to$ V5 (512tok)} & \textbf{100} & \textbf{100} & \textbf{93} & \textbf{100} & \textbf{98.3} \\
L* routing $\to$ V5 (30tok) & 100 & 100 & 12 & 100 & 78.0 \\
RPNI routing $\to$ V5 & 96 & 96 & 13 & 92 & 74.3 \\
Tool-calling agent & 95 & 76 & 5 & 59 & 58.7 \\
\hline
\multicolumn{6}{l}{\textit{V1--V5 ablation (hand-coded routing)}} \\
SLM-only baseline & 48 & 40 & 8 & 53.3 & 37.3 \\
V5 (30 tokens) & 100 & 100 & 18.7 & 100 & 79.7 \\
V5 (512 tokens) & 100 & 100 & 96 & 100 & 99.0 \\
\hline
\end{tabular}
\end{table}

\subsection{Latency Analysis}

Table~\ref{tab:latency} shows median latency by category for V5 (512 tokens) versus the best baseline. Symbolic solvers achieve three orders-of-magnitude speedups on structured categories:
\begin{itemize}
    \item AR: \textbf{1\,ms} (vs.\ 6.3\,s baseline, 6,288$\times$ speedup)
    \item ALG: \textbf{11\,ms} (vs.\ 5.8\,s baseline, 531$\times$ speedup)
    \item LOG: \textbf{2\,ms} (vs.\ 2.7\,s baseline, 1,336$\times$ speedup)
    \item WP: \textbf{137\,s} (increased from 25\,s due to higher token budget)
\end{itemize}

\begin{figure}[t]
    \centering
    \includegraphics[width=\columnwidth]{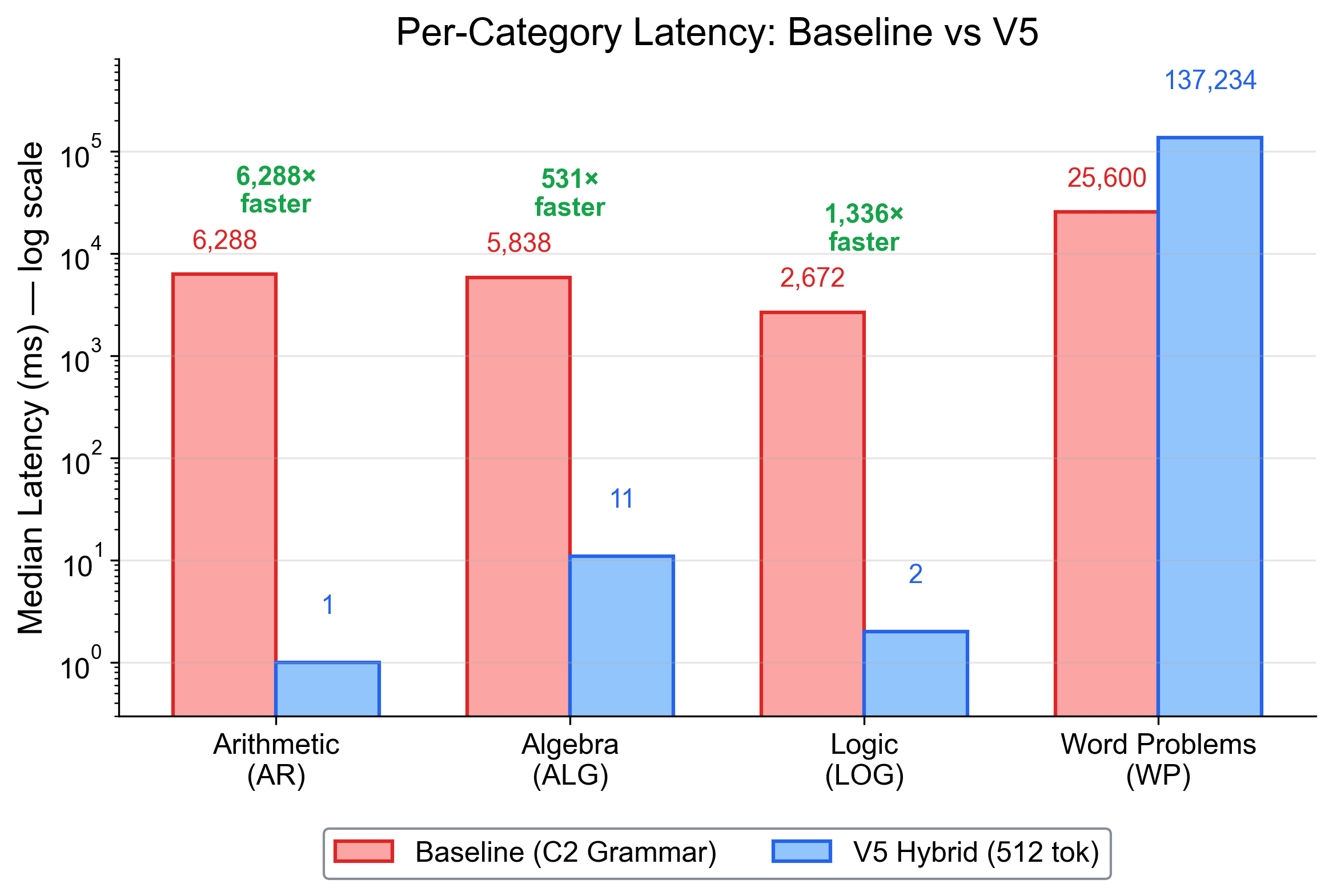}
    \caption{Per-category median latency on a log scale: baseline vs.\ V5. Symbolic solvers achieve 531--6,288$\times$ speedups on structured categories. WP latency increases due to the higher token budget.}
    \label{fig:latency}
\end{figure}

The system-wide median of 6\,ms reflects the dominance of symbolic routes (75\% of queries). However, the WP latency of approximately 137\,s per prompt is a significant practical limitation for real-time applications. Fig.~\ref{fig:latency} visualizes this on a log scale.

\begin{table}[t]
\centering
\caption{Median Latency by Category}
\label{tab:latency}
\begin{tabular}{lrrrr}
\hline
\textbf{System} & \textbf{AR (ms)} & \textbf{ALG (ms)} & \textbf{LOG (ms)} & \textbf{WP (s)} \\
\hline
Baseline (best) & 6,288 & 5,838 & 2,672 & 25.6 \\
V5 (512tok) & 1 & 11 & 2 & 137.2 \\
\hline
\end{tabular}
\end{table}

\subsection{Energy--Accuracy Tradeoff}

Table~\ref{tab:wp_ablation} shows the effect of token budget on word-problem accuracy and energy. The tradeoff is substantial: 512 tokens yields 96\% WP accuracy but costs 6.9$\times$ more energy per prompt than 30 tokens.

\begin{table}[t]
\centering
\caption{Word-Problem Accuracy vs.\ Token Budget}
\label{tab:wp_ablation}
\begin{tabular}{llcccc}
\hline
\textbf{Model} & \textbf{Tok.} & \textbf{CoT} & \textbf{WP (\%)} & \textbf{Lat.\ (s)} & \textbf{E (mWh/p)} \\
\hline
Phi-4-mini & 30 & No & 18.7 & 26.0 & 30.6 \\
Phi-4-mini & 300 & Yes & 92.0 & 128.6 & 147.2 \\
Phi-4-mini & 512 & Yes & 96.0 & 135.2 & 211.1 \\
\hline
Qwen-1.5B & 30 & No & 18.7 & 26.0 & 30.6 \\
Qwen-1.5B & 300 & Yes & 64.0 & 100.1 & 113.1 \\
Qwen-1.5B & 512 & Yes & 96.0 & 128.2 & 147.6 \\
\hline
\end{tabular}
\end{table}

\begin{figure}[t]
    \centering
    \includegraphics[width=\columnwidth]{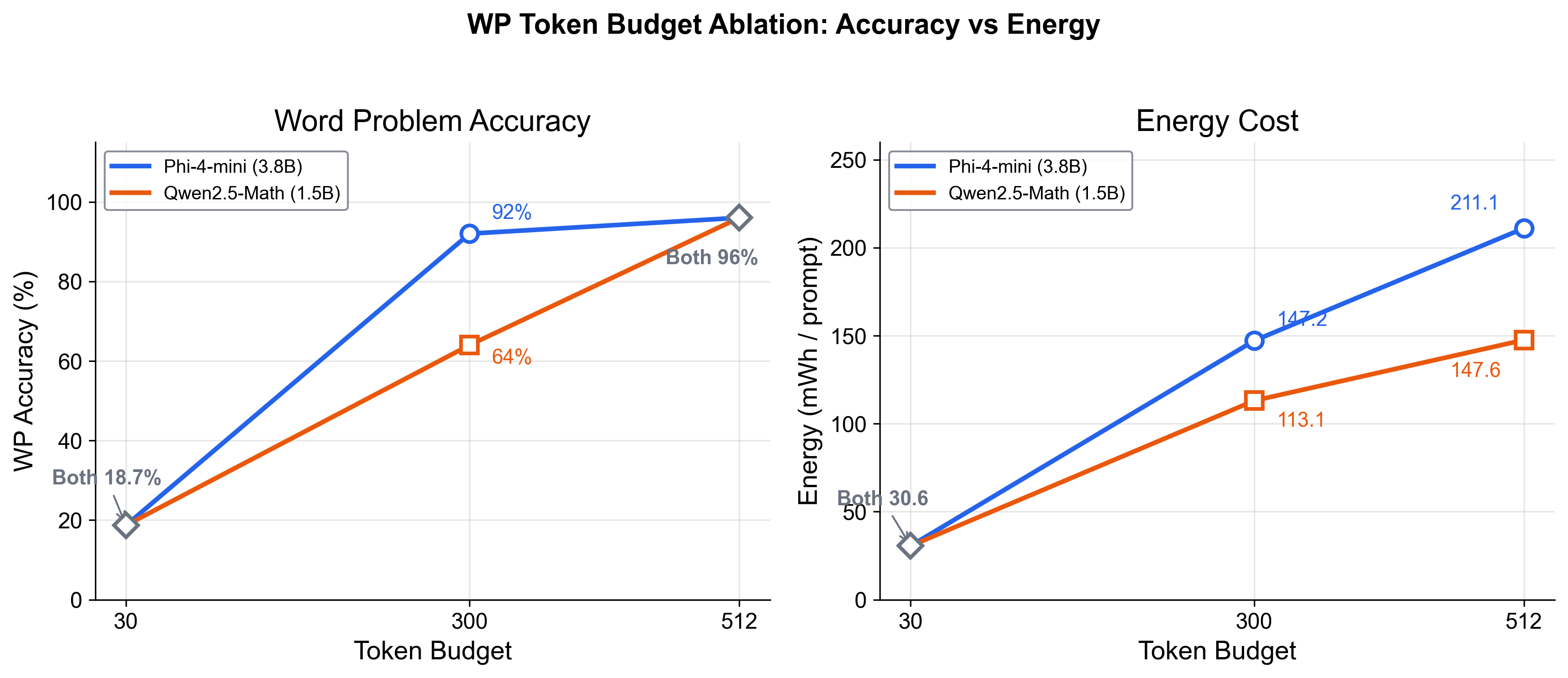}
    \caption{Word-problem token budget ablation for two SLMs: accuracy (left) and energy cost (right) as a function of token budget. Both models converge to 96\% at 512 tokens, but Phi-4-mini is more token-efficient at 300 tokens.}
    \label{fig:wp_ablation}
\end{figure}

Note that the best system (V5, Phi, 512tok) uses \textit{more} energy per prompt (211.1\,mWh) than the baseline (75.0\,mWh). The energy savings from symbolic routing on AR/ALG/LOG are outweighed by the increased token budget for WP. This tradeoff between higher accuracy and higher energy is fundamental and is discussed further in Section~\ref{sec:discussion}. 

\subsection{Error Distribution}

Table~\ref{tab:errors} shows the error taxonomy for selected systems. V5 at 512 tokens eliminates all error types except E5 (3 word-problem errors out of 300 trials), corresponding to a single prompt that fails consistently across all 3 repeats.

\begin{table}[t]
\centering
\caption{Error Distribution (300 Trials Each). E0 = Correct}
\label{tab:errors}
\begin{tabular}{lrrrrrrr}
\hline
\textbf{System} & \textbf{E0} & \textbf{E1} & \textbf{E2} & \textbf{E3} & \textbf{E5} & \textbf{E7} & \textbf{E8} \\
\hline
Baseline & 112 & 36 & 26 & 45 & 21 & 9 & 51 \\
V1 (+A5) & 159 & 0 & 27 & 45 & 24 & 0 & 45 \\
V2 (+A4) & 201 & 0 & 27 & 3 & 24 & 0 & 45 \\
V3.1 (+A3) & 212 & 0 & 27 & 0 & 18 & 30 & 13 \\
V5 (512) & 297 & 0 & 0 & 0 & 3 & 0 & 0 \\
\hline
\end{tabular}
\end{table}

% ======================================================================
\section{Analysis}
\label{sec:analysis}
% ======================================================================

\subsection{Why Symbolic Routing Works}

The most notable feature of the ablation is that the gains are not gradual but abrupt, since each solver produces a sharp jump in its own category at the moment it is introduced. A5 takes arithmetic from 48\% to 100\%, A4 takes algebra from 40\% to 96--100\%, and A6 takes logic from 64\% to 100\%. These are not the smooth improvements one expects from tuning a model. They are what happens when probabilistic approximation is replaced outright by exact computation.

\begin{figure}[t]
    \centering
    \includegraphics[width=\columnwidth]{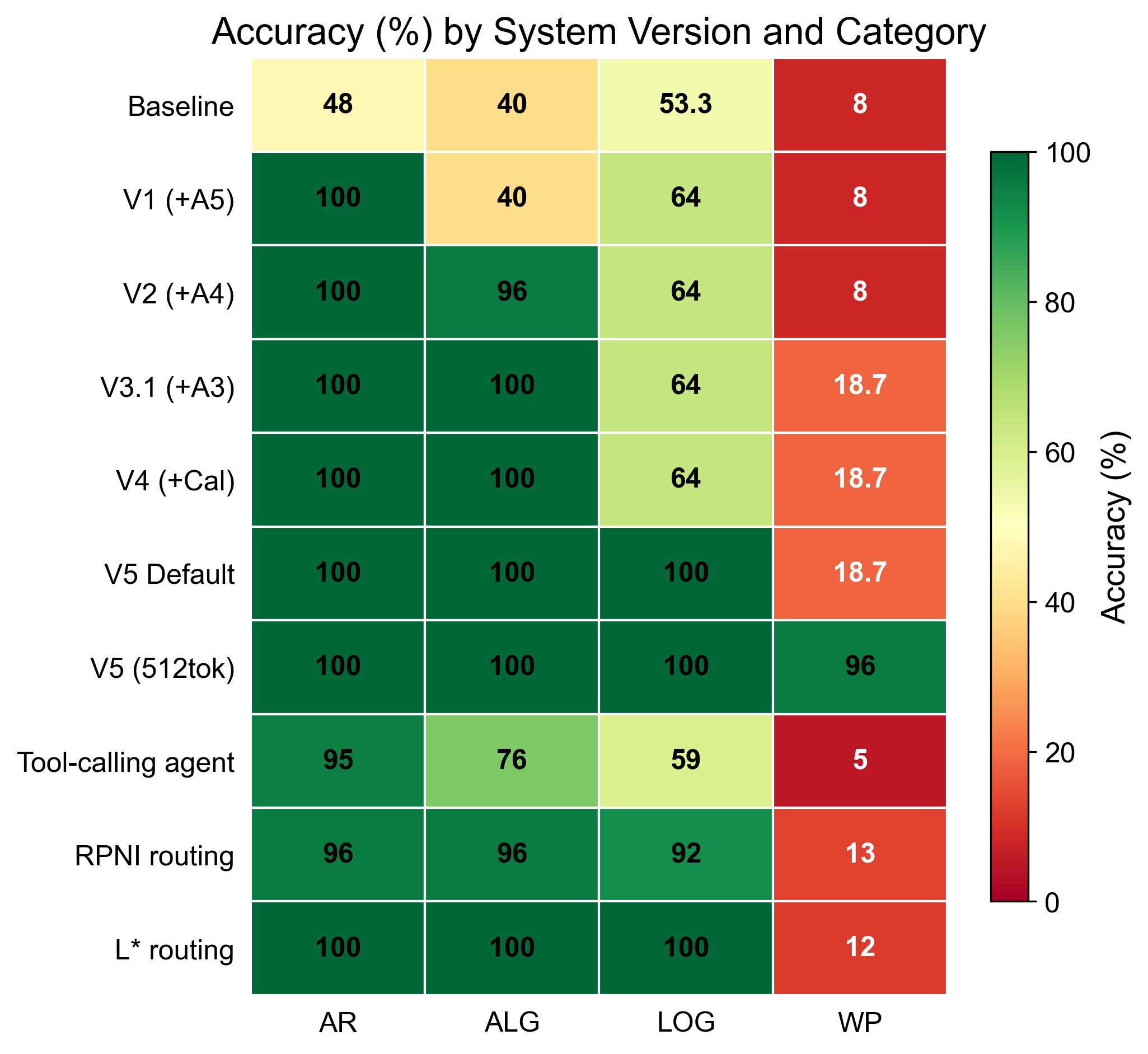}
    \caption{Accuracy by category across all system versions and configurations. Each column is a category; each row is a system version. Symbolic solvers (A5, A4, A6) produce immediate 100\% accuracy in their target categories upon introduction.}
    \label{fig:version_categories}
\end{figure}

The principle behind the architecture follows directly. Once the structure of a problem is known, a symbolic solver dominates neural inference in both accuracy and efficiency, and the SLM is needed only on inputs whose structure is not known in advance, which in this domain means word problems and little else. Fig.~\ref{fig:version_categories} shows the category-level accuracy across every version and configuration we tested.

\subsection{Error Analysis: The Single V5 Failure}
\label{sec:error_analysis}

The sole error in V5 (Phi-4-mini, 512 tokens) occurs on prompt WP\_000112 from GSM8K:

\begin{quote}
\small
\textit{``At Allan's house, there is twice as much corn as cannolis. He has a total of 40 cannolis. Allan bought 60 more cannolis at the store and 40 fewer corns than the number of cannolis. Find the combined total of the number of corns and cannolis Allan has in the house?''}
\end{quote}

Expected answer: \textbf{200}. Both models consistently produce \textbf{240}. The chain-of-thought output reveals the error:

\begin{quote}
\small
\textit{``Allan bought 60 more cannolis, so now he has $40 + 60 = 100$ cannolis. He also bought 40 fewer corns than the number of cannolis he bought, which means he bought $100 - 40 = 60$ corns.''}
\end{quote}

The model interprets ``40 fewer corns than the number of cannolis'' as referring to the \textit{updated} cannoli count (100) rather than the original count (40), yielding $100 - 40 = 60$ new corns instead of the intended $40 - 40 = 0$. This is a linguistic ambiguity in the problem statement, since the referent of ``the number of cannolis'' is genuinely unclear. Both Phi-4-mini and Qwen2.5-Math produce the same interpretation across all trials, suggesting this is a systematic parsing bias rather than a stochastic error.

\subsection{Model Comparison}

At 512 tokens, both models achieve 99.0\% overall and 96.0\% WP accuracy. However, at 300 tokens, Phi-4-mini achieves 92.0\% WP accuracy versus Qwen's 64.0\%, suggesting Phi-4-mini is more token-efficient for word-problem reasoning despite Qwen's mathematics-specific pretraining.

Both models achieve identical performance on symbolic categories (100\%) because these queries never reach the model.

% ======================================================================
\section{Discussion}
\label{sec:discussion}
% ======================================================================

\subsection{Generalizability}
Our benchmark covers four well-defined categories drawn from established datasets. Real-world queries may span additional categories (e.g., commonsense reasoning, spatial reasoning) that are neither fully symbolic nor standard word problems. However, because the routing patterns are \emph{learned} rather than hard-coded, extending to new categories requires only additional training examples and a new DFA, not manual regex engineering.

\subsection{Classifier Robustness}
\label{sec:classifier_robustness}
The L*-learned classifier achieves 100\% routing accuracy on held-out T2 prompts, but this reflects the well-separated nature of the benchmark categories rather than adversarial robustness. The token-class abstraction (Section~\ref{sec:token_abstraction}) provides some robustness to paraphrasing, since surface words are mapped to abstract types, but adversarially crafted queries could still cause misclassification. A fallback mechanism that detects solver failures and re-routes would further improve robustness.

\subsection{Active vs.\ Passive Learning}
The comparison between L* (100\% classification, 78.0\% task accuracy) and RPNI (86\% classification, 74.3\% task accuracy) demonstrates the value of the SLM membership oracle. RPNI's lower performance stems from word-problem misclassification: WP prompts have diverse linguistic structure that is difficult to capture from 10 positive examples alone. The SLM oracle allows L* to generalize beyond the training distribution by answering membership queries on novel token sequences. This suggests that even weak oracles (a 3.8B parameter model) can meaningfully improve grammatical inference when the target language has high variance.

\subsection{Dataset Size}
Our benchmark of 100 prompts (25 per category) is small by modern standards. The 95\% Wilson score interval on 99.0\% accuracy is [97.1\%, 99.8\%], which is reasonably tight, but per-category intervals are wider (for example, 96\% WP accuracy on 25 prompts has a 95\% Wilson interval of [80.5\%, 99.3\%]). Evaluation on larger subsets of the source benchmarks would strengthen the claims.

\subsection{Energy--Accuracy Tradeoff}
The best-accuracy system (V5, 512tok) uses 2.8$\times$ more energy than the baseline. The energy savings from symbolic routing on 75\% of queries are real but are outweighed by the increased WP generation cost at higher token budgets (211.1\,mWh at 512 tokens vs.\ 30.6\,mWh at 30 tokens). For battery-powered deployments, adaptive token budgets that start low and escalate only when the model's output indicates insufficient reasoning could reduce average energy while maintaining accuracy.

\subsection{Scalability of Symbolic Solvers}
Our solvers handle single-variable linear equations, basic arithmetic, and propositional logic. Extending to nonlinear equations, multi-variable systems, or first-order logic would require more powerful solvers, but the routing architecture would remain unchanged.

\subsection{Practical WP Latency}
Word problems take approximately 137\,s per prompt at 512 tokens on the Pi~4B's CPU. This is acceptable for batch processing or non-interactive applications but prohibitive for real-time use. Faster hardware (e.g., with neural accelerators) or more aggressive quantization could reduce this.

% ======================================================================
\section{Conclusion}
\label{sec:conclusion}
% ======================================================================

We set out to make a small, CPU-bound language model reliable on structured reasoning, and we found that the most effective strategy is to invoke it less often. Our neurosymbolic router learns, through the L* grammatical inference algorithm with the SLM as a membership oracle, which queries can be settled exactly by a symbolic engine and which genuinely require the model. On a Raspberry Pi~4B with no GPU, the learned router reaches 98.3\% end-to-end accuracy at a 512-token reasoning budget and 78.0\% at a 30-token budget, compared with 58.7\% for a tool-calling agent that has access to the same solvers, and at its efficient operating point it is 2.4$\times$ faster and 2.3$\times$ more energy-efficient because structured queries are dispatched in 1--11\,ms without ever invoking the SLM.

Two secondary findings sharpen the picture. Comparing L* (active, oracle-guided) against RPNI (passive, example-only) shows that even a 3.8B-parameter oracle meaningfully improves routing on linguistically diverse categories, lifting word-problem routing well beyond what positive examples alone can capture. And the five-version ablation shows that each symbolic solver contributes a clean, discontinuous gain in its target category, with no interaction effects to untangle.

The main remaining limitation is the cost of the word-problem path. Word problems are the one category that still requires the model, and reaching 93--96\% accuracy on them requires a 512-token budget that consumes roughly seven times the energy of the 30-token setting and more than two minutes of CPU time per prompt. Reducing this cost, through adaptive token budgets that increase only when a short answer appears unreliable, faster edge accelerators, and larger evaluations, is a natural direction for future work. More broadly, our results suggest that trustworthy reasoning on edge hardware depends less on larger models or cloud access than on learning to send each query to the solver that can actually answer it.

% Acknowledgment section removed for double-anonymous review.
% Will be restored in camera-ready version.

\bibliographystyle{IEEEtran}
\bibliography{references}

\end{document}